\documentclass[conference]{IEEEtran}
\IEEEoverridecommandlockouts
\usepackage{cite}
\usepackage{amsmath,amssymb,amsfonts}
\usepackage{algorithmic}
\usepackage{graphicx}
\usepackage{textcomp}
\usepackage{subcaption}
\usepackage{booktabs} % for professional tables
\usepackage{float}
\usepackage{xcolor}
\usepackage{listings}
\usepackage{cleveref}
\def\BibTeX{{\rm B\kern-.05em{\sc i\kern-.025em b}\kern-.08em
    T\kern-.1667em\lower.7ex\hbox{E}\kern-.125emX}}
\begin{document}

\title{Wildfire risk forecast: \\An optimizable fire danger index}

\author{\IEEEauthorblockN{Eduardo Rodrigues}
\IEEEauthorblockA{\textit{IBM Research, Brazil}}
edrodri@br.ibm.com
\and
\IEEEauthorblockN{Bianca Zadrozny}
\IEEEauthorblockA{\textit{IBM Research, Brazil}}
biancaz@br.ibm.com
\and
\IEEEauthorblockN{Campbell D. Watson}
\IEEEauthorblockA{\textit{IBM Research, USA}}
cwatson@us.ibm.com
}

\maketitle

\begin{abstract}
    Wildfire events have caused severe losses in many places around the world and 
    are expected to increase with climate change. Throughout the years many technologies
    have been developed to identify fire events early on and to simulate fire behavior
    once they have started. Another particularly helpful technology is fire risk indices,
    which use weather forcing to make advanced predictions of the risk of fire. Predictions of fire risk indices can
    be used, for instance, to allocate resources in places with high risk. These indices have been developed 
    over the years as empirical models with parameters that were estimated in lab 
    experiments and field tests. These parameters, however, may not fit well all places
    where these models are used. In this paper we propose a novel implementation of
    one index (NFDRS IC) as a differentiable function in which one can optimize its
    internal parameters via gradient descent. We leverage existing machine learning
    frameworks (PyTorch) to construct our model. This approach has two benefits: (1)
    the NFDRS IC parameters can be improved for each region using actual observed 
    fire events, and (2) the internal variables remain intact for interpretations by specialists instead of meaningless hidden layers as in traditional 
    neural networks. In this paper we evaluate our strategy with actual fire events for locations in the USA and Europe.
\end{abstract}


\section{Introduction}

Wildfires have caused billion-dollar disasters \cite{ncei2020noaa} and
taken the lives of many people \cite{Bateman2018}. The threat of wildfires has been exacerbated by climate change through more frequent droughts and longer wildfire seasons. Land use change has also played a role. As an example, the number of large fires in the western United States has doubled from 1984 to 2015 \cite{Droughts}. An important tool in tackling this problem 
is wildfire risk index models. Such models are typically driven by atmospheric data (e.g., precipitation, wind, humidity)
and are used as an early warning system for people to be evacuated and preventative action taken (e.g. the trimming trees close to power lines). 

Over the years, several fire index systems have been developed which take a range of 
input data, e.g. temperature, relative humidity, fuel type, vegetation stage, topography etc, and output
indices related to fire potential. Examples of index systems are the Canadian Forest Service
Fire Weather Index Rating System (FWI) \cite{van1974structure}, the Australian McArthur rating
systems (Mark 5) \cite{mcarthur1966weather} and National Fire-Danger Rating System (NFDRS) \cite{deeming1977national}.
These index systems are based on both physical and empirical conditions and, to this date, are used to generate
risk maps which inform agencies, individuals and companies the risk in their particular areas \cite{effis}.

Internally, these fire index models compute intermediate variables which are combined to produce a set of fire danger indices. These internal variables are meaningful in that they correspond to (possibly) measurable quantities; e.g., the equilibrium moisture content, which represents the steady state moisture content of woody material, and the slope effect coefficient, among many others. The relations among the internal variables and the indices are established empirically, with parameters that have been estimated throughout the years. These parameters, however, may not be the most appropriate for all regions where these indices are applied. Our hypothesis in this short paper is that we can adjust the internal parameters to best fit the particular regions in which one intends to use the wildfire risk indices. This approach has two major advantages: (1) one can train the index models using actual observed fire, but starting with an already proven index, and (2) it preserves the internal variables which are meaningful for wildfire specialists.

In order to test our hypothesis we implemented the \textit{ignition component} (IC) index of the National Fire Danger Rating System (NFDRS) as a smooth function so that we can apply stochastic gradient descent (SGD) and optimize the internal parameters. This approach is similar to a neural network though with the difference that it is not a traditional architecture but a smooth version of a fire index model. In this paper, we present results comparing the unmodified model against a trained model (trained with observed fire from 2010 to 2015) for the period 2016 to 2020 in three separate regions.

\section{Model}

Our model is based on the U.S. Forest Service - National Fire-Danger Rating System (NFDRS). In particular, we compute the \textit{ignition component} (IC). This index gives a numerical rating of the probability that a fire (which requires suppression action) will result if a firebrand is introduced into a fine fuel complex. This index has been shown to indicate fire danger \cite{di2016potential} and is commonly  used throughout the United States. There are other indices in the NFDRS system which can be made optimizable, however in this report we focus only on IC.

\begin{figure*}
  \begin{center}
  \includegraphics[width=.85\textwidth]{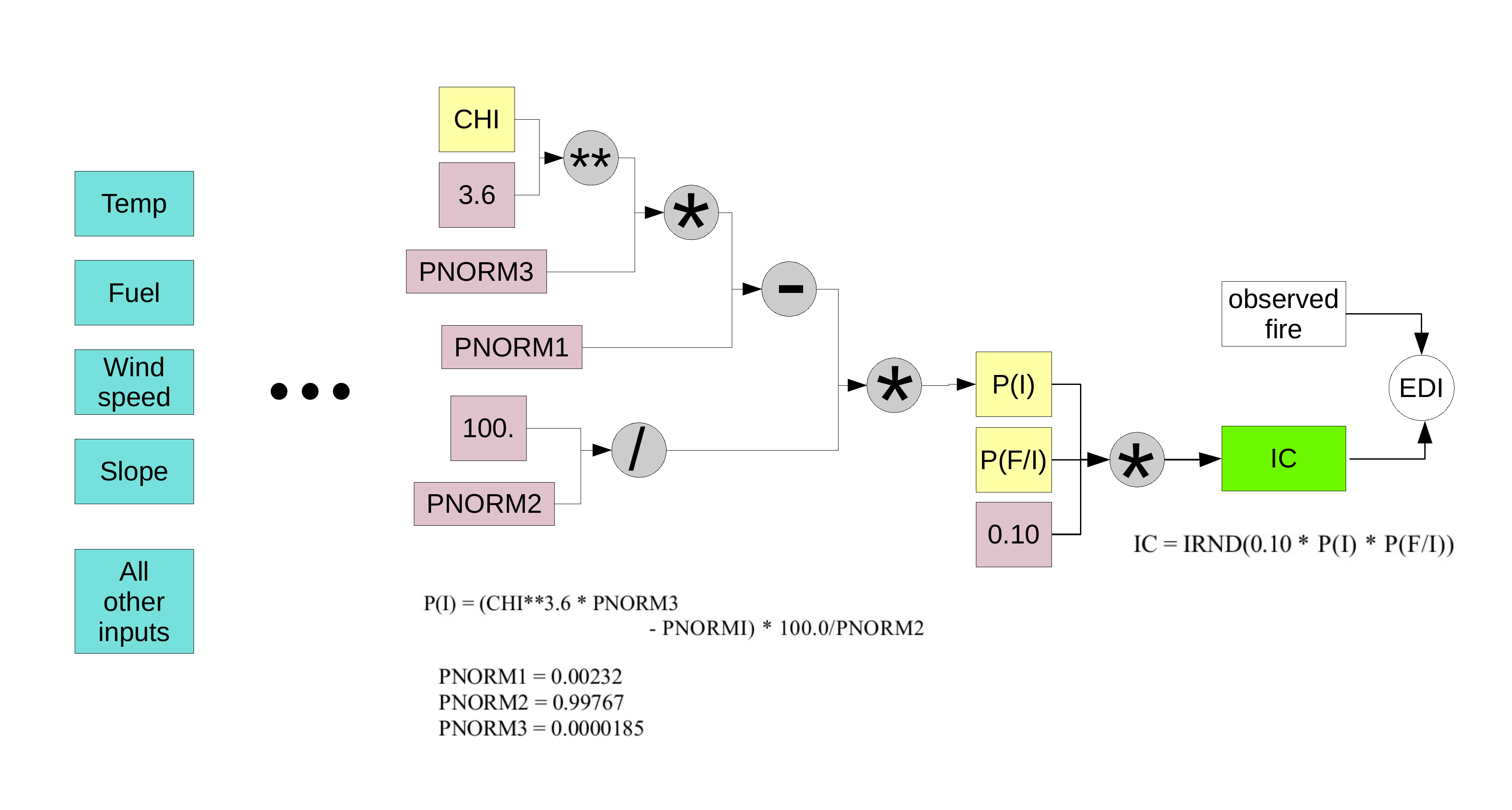}
  \end{center}
  \caption{Head of the model. Yellow boxes are intermediate variables, pink boxes are parameters, light blue boxes are inputs, and operations are in grey. The Extremal Dependency Index (EDI) is the optimization criterion (loss function).}
  \label{fig0}
\end{figure*}

In order to produce the IC index, a number of input variables is used. These are: temperature, maximum temperature, minimum temperature, relative humidity, maximum relative humidity, minimum relative humidity (the last two may be estimated as described in the report \cite{cohen1985national}), wind speed, cloud cover, precipitation duration (which could be estimated as in \cite{cohen1985national}), mean cumulative annual precipitation (climatological precipitation mean), vegetation stage (four classes: (1) cured, (2) pre-green, (3) green and (4) transition), vegetation cover (derived from GLCC as described in \cite{di2016potential}), slope (five classes as described in the report \cite{cohen1985national}), fuel model (NFDRS Fuel Model Map for the USA and extending globally as in \cite{di2016potential}), and climatic zones (Köppen classification).

Many \textit{intermediate variables} need to be computed out of the input variables before IC is produced. Most of these intermediate variables have concrete meaning and could be measured. For example, 1-hour, 10-hour, 100-hour, and 1000-hour dead-fuels moisture (which represent wildland fuels whose moisture contents are controlled entirely by changing weather conditions), live-fuel moisture variables, reaction velocity, slope effect, and many others. These intermediate variables, and the final IC index itself, were estimated in laboratory and field experiments and their relations are embedded through specific \textit{parameters} in the NFDRS model. The head of the model can be seen in Figure \ref{fig0} (all relations up to inputs can be found in \cite{cohen1985national}). 

Our hypothesis in this paper is that we can optimize the internal parameters of the IC index so that it better predicts fire danger for particular regions, but keeps the internal variables intact. The major benefit of this approach is that one can still interpret the internal variables and possibly measure them. In order to do that, we need an optimization criterion and procedure to adjust parameters.

There are many possible optimization criteria (aka loss functions). All of them will compare the prediction (in our case the IC index) with a measure of fire danger and assign a score. However, it is not easy to obtain an accurate measure of fire danger. Therefore, one needs a proxy that may not be perfect but correlates with wildfire risk. Observed fire is obviously the most appropriate proxy, but one needs to keep in mind that a region may experience a high risk of wildfire though no wildfire eventuates. Moreover, fire events are extremely rare compared to non-fire events in a large area over time. This leads to a very unbalanced scenario in which a trivial solution (that is, no-fire ever) will perform statistically very well depending on the optimization criterion. Consequently, a loss function needs to be less dependent on the base rate. We therefore use the extremal dependency index (EDI) \cite{ferro2011extremal}:

\begin{equation*}
    \text{EDI} = \frac{\text{log}F-\text{log}H}{\text{log}F+\text{log}H}
\end{equation*}

\noindent (where $\text{F}$ is the false-alarm rate and $\text{H}$ is the hit rate), that is because it is a non-vanishing measure for rare events which also have been used for fire events. EDI provides a score in the range $[-1, 1]$, in which 1 is for a perfect forecast and 0 is for a random forecast.

\begin{figure}[htbp]
\centerline{\includegraphics[width=0.4\textwidth]{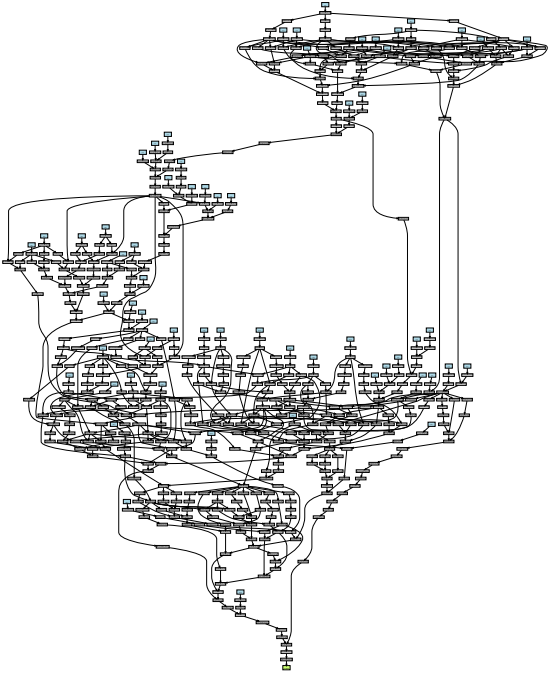}}
\caption{Network architecture illustrating the depth of the model. Green box is the output (IC index), grey boxes are operations, and light blue boxes are inputs. Intermediate variables and parameters are not shown.}
\label{fig1}
\end{figure}

As for the procedure to adjust parameters, we use gradient descent and use a validation procedure to avoid overfitting (the details are described in the next section). In order to use gradient descent, we implemented the IC index as a differentiable function with PyTorch \cite{paszke2019pytorch}. In addition to the sheer size of the model (see Figure \ref{fig1} where blue boxes are inputs, gray boxes are operations and the green box is the IC index output), implementing the IC computations as a differentiable function poses a few challenges. The first one is that there are many hard branches in the model which makes the function non-differentiable. For example, IC is set to zero if the expression:

\begin{equation*}
    \left((344.-\text{QIGN})/10.\right)^{3.6}*\text{PNORM3}
\end{equation*}

\noindent is equal to or less than $\text{PNORM1}$. Where $\text{QIGN}$ is an intermediate variable that means \textit{heat of ignition}, $\text{PNORM1}$ and $\text{PNORM3}$ are scaling factors. This makes IC non-differentiable with respect to that condition. In order to make it differentiable, we had to implement the following smooth pattern of this branch (and similar ones throughout the model):

\begin{lstlisting}
  if X < A then
    return Y
  else
    return Z
\end{lstlisting}

\noindent which is translated as:

\begin{equation*}
    \sigma\left((X-A)*\alpha\right) * (Z - Y) + Y
\end{equation*}

\noindent where $\sigma(x)$ is the sigmoid function and $\alpha$ is a (learnable) shape parameter.

The second challenge to make the IC index differentiable is related to where the derivative is evaluated. The internal operations may be differentiable, but the point where they are evaluated can be undefined or infinity, and this causes the training procedure to fail. Here we show a couple examples.

An internal variable in the NFDRS model is $P(F/I)$, which stands for the probability of a reportable fire and is defined as:

\begin{equation*}
    \text{P}(\text{F}/\text{I}) = \sqrt{\text{SCN}}
\end{equation*}

\noindent where $\text{SCN}$ is the normalized rate of spread. Consider a weight $w$ downstream in the model. In order to compute an update to $w$ so as to minimize the loss $\mathcal{L}$, one would need to compute the derivative of $\text{P}(\text{F}/\text{I})$ with respect to $\text{SCN}$:

\begin{equation*}
    \frac{\partial \mathcal{L}}{\partial w} = \frac{\partial \mathcal{L}}{\partial \text{P}(\text{F}/\text{I})} \frac{\partial \text{P}(\text{F}/\text{I})}{\partial \text{SCN}}\frac{\partial \text{SCN}}{\partial w}
\end{equation*}

\noindent which exists but is infinite when $\text{SCN}$ is zero, but physically it can indeed be zero. To avoid the derivative blowing up we clip the gradients.

Another example which has a derivative, but it is not defined at a point is in the definition of $\text{QIGN}$:

\begin{equation*}
\begin{split}
\text{QIGN} =\ 144.5 - (0.266 * \text{TMPPRM})\\
  - (0.00058* \text{TMPPRM}^{2.0})\\
  - (0.01 * \text{TMPPRM} * \text{MC1})\\
  + (18.54 * (1.0 - \text{EXP}(-0.151 * \text{MC1})))\\
  + 6.4 * \text{MC1})\\
\end{split}
\end{equation*}

This definition has many potential candidates for parameters. However, the square of $\text{TMPPRM}$ cannot be made into a parameter because its derivative is not defined for negative values of $\text{TMPPRM}$ but in practice $\text{TMPPRM}$ can be negative because it is the temperature estimated for fuel-atmosphere interface in Fahrenheit. In this case, the work around is simply to not make the square into a parameter. The model has similar characteristics in other places. 

In the next section, we show the procedure to train and test our model using 10 years of data for three different locations.

\section{Evaluation}

For the purpose of evaluating our model, we compared the original IC index with a trained model. The training data is daily and goes from 2010 to 2015, and the testing data, also daily, from 2016 to 2020. We computed EDI for California, Texas and Italy in the testing set to make our comparison.

Weather input data (temperature, RH, wind speed, cloud cover and precipitation) comes from the ERA5 reanalysis dataset. We did not use forecasted weather data so as to avoid errors in the forecast that would impact the performance of the index (both trained and untrained). Consequently these results represent the potential predictability of wildfire. Climatic zones were obtained from \cite{koppen}, the USA fuel map from \cite{fuelmap} and the  Europe fuel map \cite{fuelmapeurope}, slope was obtained from \cite{gtopo30}, vegetation cover is built from the GLCC dataset \cite{loveland2000development}, and vegetation stage classes are derived from \cite{lai}. Finally, fire observations are obtained from the MODIS/Terra+Aqua Burned Area Monthly L3 product. 

All datasets were placed in the same resolution of 25km over the same grid points. This resolution is the same as the results presented in \cite{di2016potential}. Continuous variables were interpolated linearly, while classes were interpolated by nearest neighbors.

In order to evaluate the skill, the observed fire dataset were pooled into the same resolution as the input, i.e. if at least one fire event is observed in the original resolution, the resulting grid point in the target resolution will record a fire. In addition, a fire prediction in a neighboring grid point where fire did not occur counts as hit ("fuzzy" pixel \cite{di2016potential}).

A fire is forecast when the IC index value is above the second quantile of the index distribution, which has been previously computed from the training set. This forecast along with the observed fire dataset is fed into the EDI function for evaluation. 

\begin{figure}[h!]
\centering
\begin{subfigure}[b]{0.47\textwidth}
   \includegraphics[width=\textwidth]{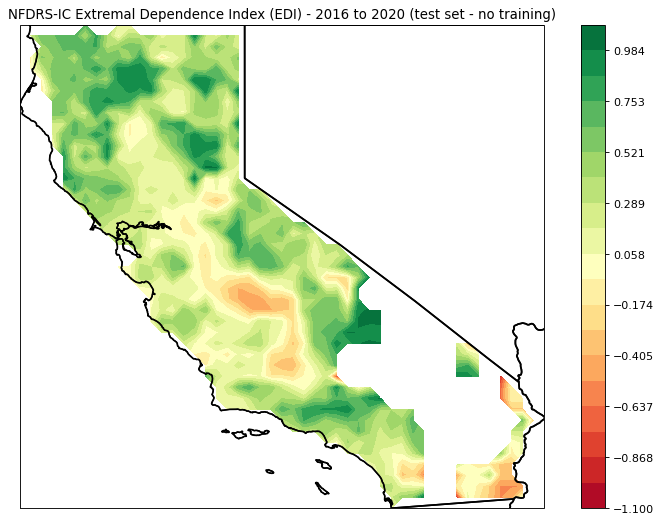}
   \caption{Untrained model}
   \label{fig:res11}
\end{subfigure}
\begin{subfigure}[b]{0.47\textwidth}
   \includegraphics[width=\textwidth]{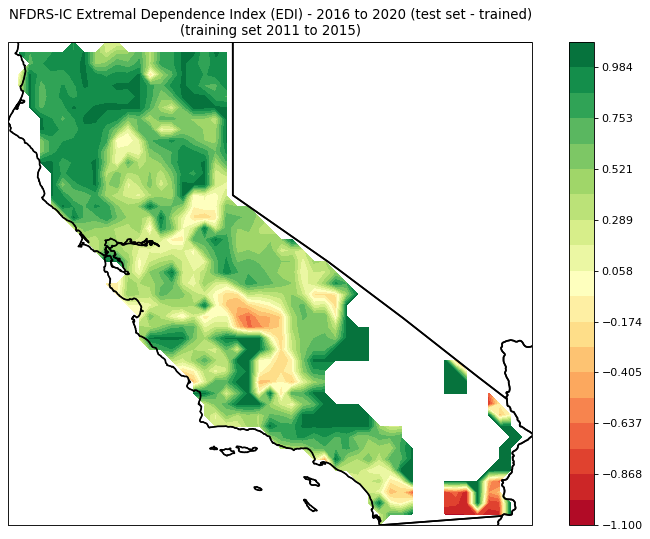}
   \vspace{-0.7cm}
   \caption{Trained model}
   \label{fig:res12}
\end{subfigure}
\begin{subfigure}[b]{0.47\textwidth}
   \includegraphics[width=\textwidth]{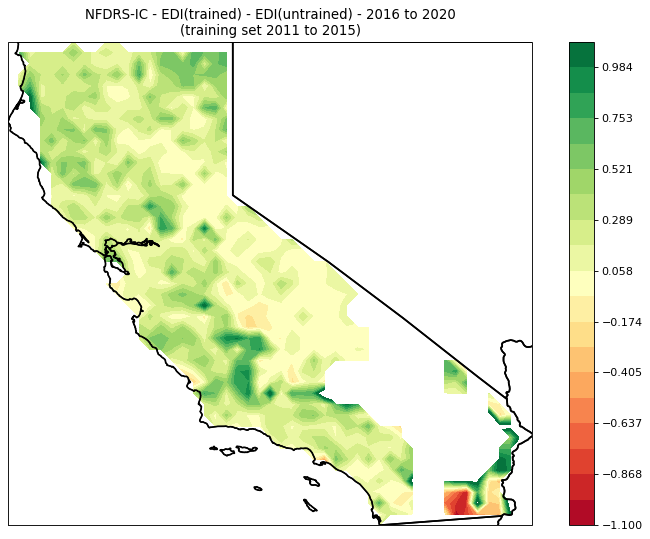}
   \vspace{-0.7cm}
   \caption{Differnce between trained EDI and untrained EDI}
   \label{fig:res13}
\end{subfigure}
   \caption{Evaluation of IC for California over the period from 2016 to 2020 (training set from 2010 to 2015)}
\label{fig:res1}
\end{figure}

\begin{figure}[h!]
\centering
\begin{subfigure}[b]{0.47\textwidth}
   \includegraphics[width=\textwidth]{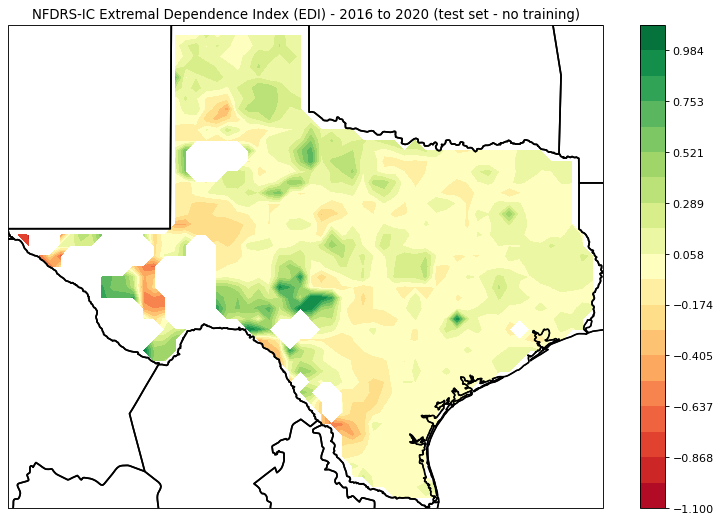}
   \caption{Untrained model}
   \label{fig:res21}
\end{subfigure}
\begin{subfigure}[b]{0.47\textwidth}
   \includegraphics[width=\textwidth]{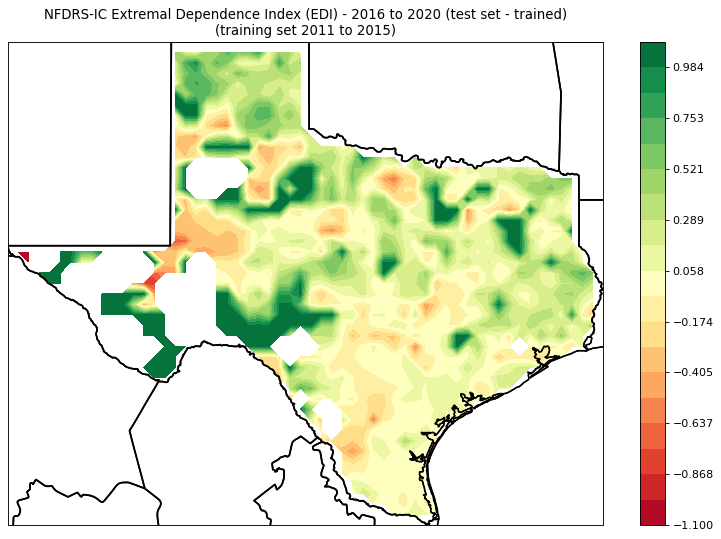}
   \vspace{-0.7cm}
   \caption{Trained model}
   \label{fig:res22}
\end{subfigure}
\begin{subfigure}[b]{0.47\textwidth}
   \includegraphics[width=\textwidth]{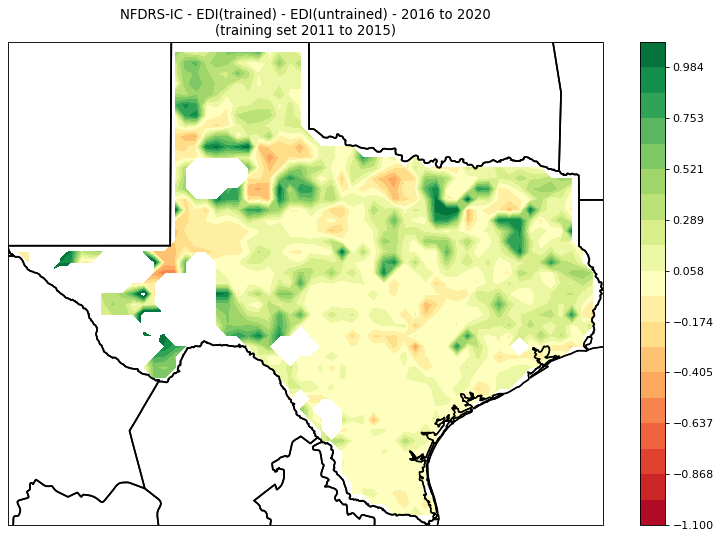}
   \vspace{-0.7cm}
   \caption{Differnce between trained EDI and untrained EDI}
   \label{fig:res23}
\end{subfigure}
   \caption{Evaluation of IC for Texas over the period from 2016 to 2020 (training set from 2010 to 2015)}
\label{fig:res2}
\end{figure}

\begin{figure}[h!]
\centering
\begin{subfigure}[b]{0.47\textwidth}
   \includegraphics[width=\textwidth]{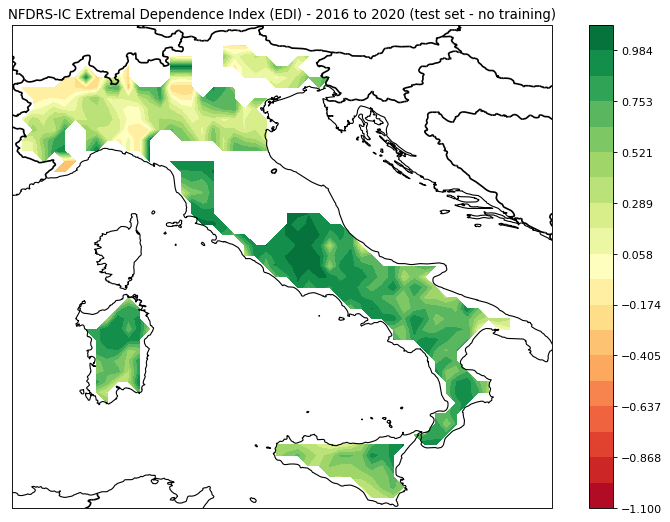}
   \caption{Untrained model}
   \label{fig:res31}
\end{subfigure}
\begin{subfigure}[b]{0.47\textwidth}
   \includegraphics[width=\textwidth]{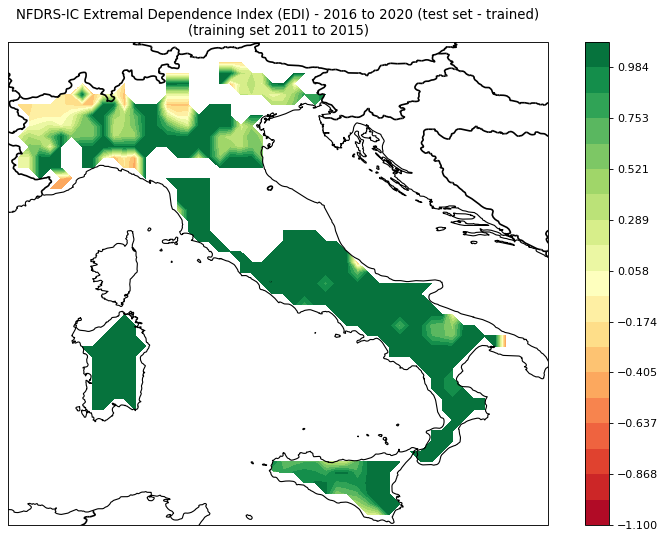}
   \vspace{-0.7cm}
   \caption{Trained model}
   \label{fig:res32}
\end{subfigure}
\begin{subfigure}[b]{0.47\textwidth}
   \includegraphics[width=\textwidth]{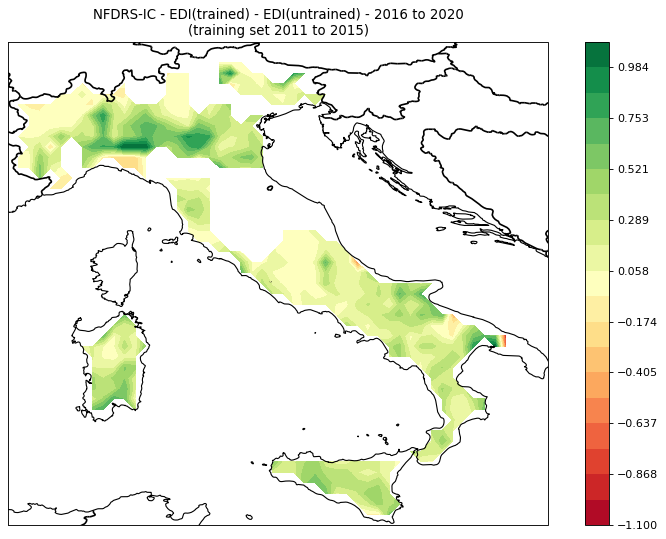}
   \vspace{-0.7cm}
   \caption{Differnce between trained EDI and untrained EDI}
   \label{fig:res33}
\end{subfigure}
   \caption{Evaluation of IC for Italy over the period from 2016 to 2020 (training set from 2010 to 2015)}
\label{fig:res3}
\end{figure}

Results can be seen in \cref{fig:res1,fig:res2,fig:res3} for California, Texas and Italy respectively. The first sub-figure in that sequence represents EDI for the untrained model (the default NFDRS IC index), the second sub-figure EDI for the trained model (the optimized NFDRS IC index), and the last sub-figure the difference between trained and untrained. EDI can take values from 1 to -1, and one representing perfect forecast and 0 for random forecast. 

Overall one can see improvements of the model skill over most of the areas in the testing set. Particularly, for California in Sierra Nevada and Central Valley, the skill improves compared to the baseline. At the extreme south, however, the trained model performs worst. This region, nevertheless, has few fire events in the testing set (now shown), so we hypothesize this is due to outliers. The Italy map shows similar behavior in which most places show improvements. Texas has more mixed results. The original (untrained) model is close to random, with just a few good places mainly in the Edwards plateau. The trained model improves skill in sparse areas in the north but at the cost of some other areas also in the north. 

One possible explanation for the mixed results over Texas is the input data. Fuel maps have a large impact on the IC index as well as vegetation stage. Our model can in principle be used to optimize input maps by using gradient descent all the way up to the inputs and adjusting them as if they were parameters. We are not exploring this idea in this paper however.

\section{Final remarks}

Fire indices have been used worldwide to tackle wildfire. These models have been developed over many years as a mixture of physical and empirical models. Internally, they have many meaningful variables which can be measured and interpreted by experts. 

In this paper, we presented a new approach for fire indices. Our hypothesis is that the parameters that relate internal variables and indices may not be the most appropriate for all regions. Consequently, one can find better parameters in the space of all possible values. 

In order to search the space of parameters for specific regions, we propose to recast an existing model as a differentiable function, similar to a neural network. However, instead of a meaningless hidden layer, our model has the same internal variables as the original model we implemented. Consequently, our index preserves the meaning of the internal variables.

To evaluate our model, we ran experiments over three separate regions: California, Texas and Italy. Skill improved in most places when using the trained model (i.e., the optimized NFDRS IC index). In addition, the resulting parameters differ among the regions indicating they adjusted to the particularities of the specific location. We intend to evaluate with experts the meaning of the adjustments found by the optimization procedure. We also intend to explore constraints to the parameters, so that the range in which the internal parameters vary does not go beyond what is coherent. For this, however, we will also need expert knowledge. Altogether, our strategy is intended to be used in close partnership with experts, enhancing their ability to explore changes to the model, but relying on them to provide meaning. 

%    \item Possible extension: consider neighbor cells (convolution)

%In order to implement the IC index one could take as input the IC index itself, additional inputs and then fit a traditional machine learning method. With an appropriate evaluation metric (loss function), this approach can improve the index for specific regions based on actual fire events. However, this approach

\bibliographystyle{plain}
\bibliography{ref}

\end{document}